\documentclass[sigconf]{acmart}
\AtBeginDocument{%
  }
\usepackage{booktabs}
\usepackage{multirow}
\usepackage[utf8]{inputenc}
\usepackage{amsmath}
\usepackage{algorithm}
\usepackage{algorithmic}
\usepackage{diagbox}
\usepackage{bbm}
\setcopyright{cc}
\setcctype{by-nc-nd}
\copyrightyear{2026}
\acmYear{2026}
\acmDOI{10.1145/3770855.3817817}
\acmConference[KDD '26]{Proceedings of the 32nd ACM SIGKDD Conference on Knowledge Discovery and Data Mining V.2}{August 09--13, 2026}{Jeju Island, Republic of Korea}
\acmBooktitle{Proceedings of the 32nd ACM SIGKDD Conference on Knowledge Discovery and Data Mining V.2 (KDD '26), August 09--13, 2026, Jeju Island, Republic of Korea}
\acmISBN{979-8-4007-2259-2/2026/08}
\begin{document}

\title{EPTS: Elastic Post-Training Sparsity for Efficient Large Language Model Compression}


\author{Ke Xu}
\authornote{Also affiliated with the State Key Laboratory of Opto-Electronic Information Acquisition and Protection Technology.}

\email{xuke@ahu.edu.cn}
\affiliation{%
  \department{School of Artificial Intelligence}
  \institution{Anhui University}
  \city{Hefei}
  \state{Anhui}
  \country{China}
}

\author{Jiaqi Wan}
\email{jqwan@stu.ahu.edu.cn}
\affiliation{%
  \department{School of Artificial Intelligence}
  \institution{Anhui University}
  \city{Hefei}
  \state{Anhui}
  \country{China}
}

\author{Wenhao Hu}
\email{wa2314071@stu.ahu.edu.cn}
\affiliation{%
  \department{School of Artificial Intelligence}
  \institution{Anhui University}
  \city{Hefei}
  \state{Anhui}
  \country{China}
}

\author{Han Pu}
\email{puhan@tiangong.edu.cn}
\affiliation{%
\department{School of Computer Science and Technology}
  \institution{Tiangong University}
  \city{Tianjin}
  \country{China}
}

\author{Xiaoyun Wang}
\authornote{Corresponding author.}
\email{16120304@bjtu.edu.cn}
\affiliation{%
  \department{School of Artificial Intelligence}
  \institution{Anhui University}
  \city{Hefei}
  \state{Anhui}
  \country{China}
}

\renewcommand{\shortauthors}{Ke Xu, Jiaqi Wan, Wenhao Hu, Han Pu, \&Xiaoyun Wang}

\begin{abstract}
  Post-Training Sparsity (PTS) has emerged as a crucial paradigm for compressing Large Language Models to facilitate efficient deployment on resource-constrained devices. However, existing PTS methodologies are typically confined to Single-Sparsity optimization, necessitating a separate, time-consuming optimization session for each specific sparsity level. This rigid paradigm significantly hinders flexible deployment across diverse hardware scenarios, as adapting to a new sparsity requirement mandates a complete re-optimization process. To address these limitations, we propose Elastic Post-Training Sparsity (EPTS), a unified Multi-Sparsity framework that produces a single elastic model capable of maintaining robust performance across diverse sparsity configurations through a one-shot optimization process. Specifically, we design a Multi-Sparsity Hierarchy LoRA (MS-HiLoRA) mechanism that facilitates knowledge inheritance from low- to high-sparsity groups, effectively mitigating the competition for parameter reconstruction. Furthermore, we introduce a Multi-Sparsity Feature Mixer (MSFM), which significantly enhances the model's adaptability to pruning perturbations by dynamically fusing feature representations of varying sparsity granularities. Extensive experiments on LLaMA and OPT families demonstrate that EPTS achieves competitive performance compared to state-of-the-art methods like SparseGPT and Wanda, while offering significant efficiency gains by enabling multi-scenario deployment from a single optimization. our source code is available at \url{https://github.com/xuke225/EPTS}.
\end{abstract}




\maketitle
\newcommand\kddavailabilityurl{https://doi.org/10.5281/zenodo.20371910}
\ifdefempty{\kddavailabilityurl}{}{
\begingroup\small\noindent\raggedright\textbf{Resource Availability:}\\
The source code of this paper has been made publicly available at \url{\kddavailabilityurl} and \url{https://github.com/xuke225/EPTS}.
\endgroup
}
\section{Introduction}
Large language models(LLMs)~\cite{Touvron2023Llama2O,Achiam2023GPT4TR} have demonstrated exceptional performance across numerous core domains such as computer vision, natural language processing, and information retrieval. Particularly in the field of Natural Language Processing (NLP), the rise of LLMs has achieved groundbreaking progress, with performance in tasks such as text generation, machine translation, and question-answering systems approaching or even surpassing human levels. This has laid a solid foundation for technological innovation in related fields. However, the improvement in LLMs performance often comes with a dramatic expansion in model scale. Notably, the parameter count of LLMs has surged from the millions in early periods to the current hundreds of billions. This directly leads to significant computational overhead and memory demands during model inference. Such resource-intensive characteristics pose severe challenges for deploying LLMs on resource-constrained edge devices, creating a critical bottleneck that hinders their practical application. To overcome these challenges, extensive research~\cite{DBLP:conf/aaai/XuM23} has been conducted focusing on model compression techniques, primarily including model quantization\cite{QLLM,Du-quant}, network sparsity~\cite{SparseGPT,Wanda,Dual-Asses,eraser} and knowledge distillation~\cite{Knowledge-distillation}.

As one of the core techniques in model compression, Network Pruning~\cite{DBLP:conf/nips/CunDS89,DBLP:conf/icnn/HassibiSW93,hansong} removes redundant weights that contribute minimally to model performance, constructing a sparse model to reduce computational redundancy. Traditional pruning methods primarily rely on end-to-end Retraining~\cite{DBLP:conf/iclr/LiuSZHD19,DBLP:conf/mlsys/BlalockOFG20} or training from scratch~\cite{lottery}, which not only necessitates a complete training dataset but also incurs substantial computational resource consumption and prolonged training cycles. These requirements significantly hinder their applicability in scenarios requiring rapid deployment or involving sensitive data privacy. To surmount these limitations, Post-Training Sparsity has emerged as a resource-friendly paradigm. Its core advantage lies in eliminating the computationally expensive retraining process by relying on a small amount of calibration data to preserve performance. To maximize recovery under such data-constrained conditions, existing methodologies have evolved primarily along two dimensions: high-precision weight reconstruction and optimal sparsity allocation. In terms of {weight reconstruction}, advanced approaches leverage second-order statistics, such as Hessian-based approximations~\cite{OBC} or Fisher information~\cite{WoodFisher,AFastPTP}, to accurately calibrate the remaining weights, thereby minimizing the quantization error of the loss landscape. In parallel, regarding {sparsity allocation}, the {Post-Training Sparsity (PTS)} paradigm, pioneered by~\cite{POT} via heuristic layer-wise calibration, has evolved towards unified optimization frameworks~\cite{FCPTS,UniPTS}, which treat sparsity allocation as a global search problem, automatically determining the optimal layer-wise sparsity ratios under specific latency or accuracy constraints. However, these methods require a separate optimization process for each specific sparsity. Taking UniPTS as an example, obtaining a sparse model for a specific sparsity takes several hours, and the cumulative time required for multiple sparsity rates becomes highly time-consuming. 

In the realm of LLM compression, the focus has shifted towards highly efficient paradigms. SparseGPT~\cite{SparseGPT} represents the training-free category, utilizing second-order statistics for weight compensation. Furthermore, methods like Wanda~\cite{Wanda} and RIA~\cite{RIA} adopt an optimization-free paradigm, bypassing weight updating by relying solely on novel activation-aware metrics for rapid pruning. Nevertheless, these methods share a common limitation: performance degradation at high sparsity levels. To address this, ICP~\cite{ICP} improves performance at mid-to-high sparsity levels without full-model fine-tuning or increasing peak memory overhead. It achieves this by rearranging predefined sparsity levels and employing a sliding-window-based block-level compensation pruning strategy. However, ICP remains constrained by the fact that a single training session yields a sparse model for only one specific sparsity rate.
\begin{figure*}[t!]   
    \centering
    \includegraphics[width=0.96\linewidth]{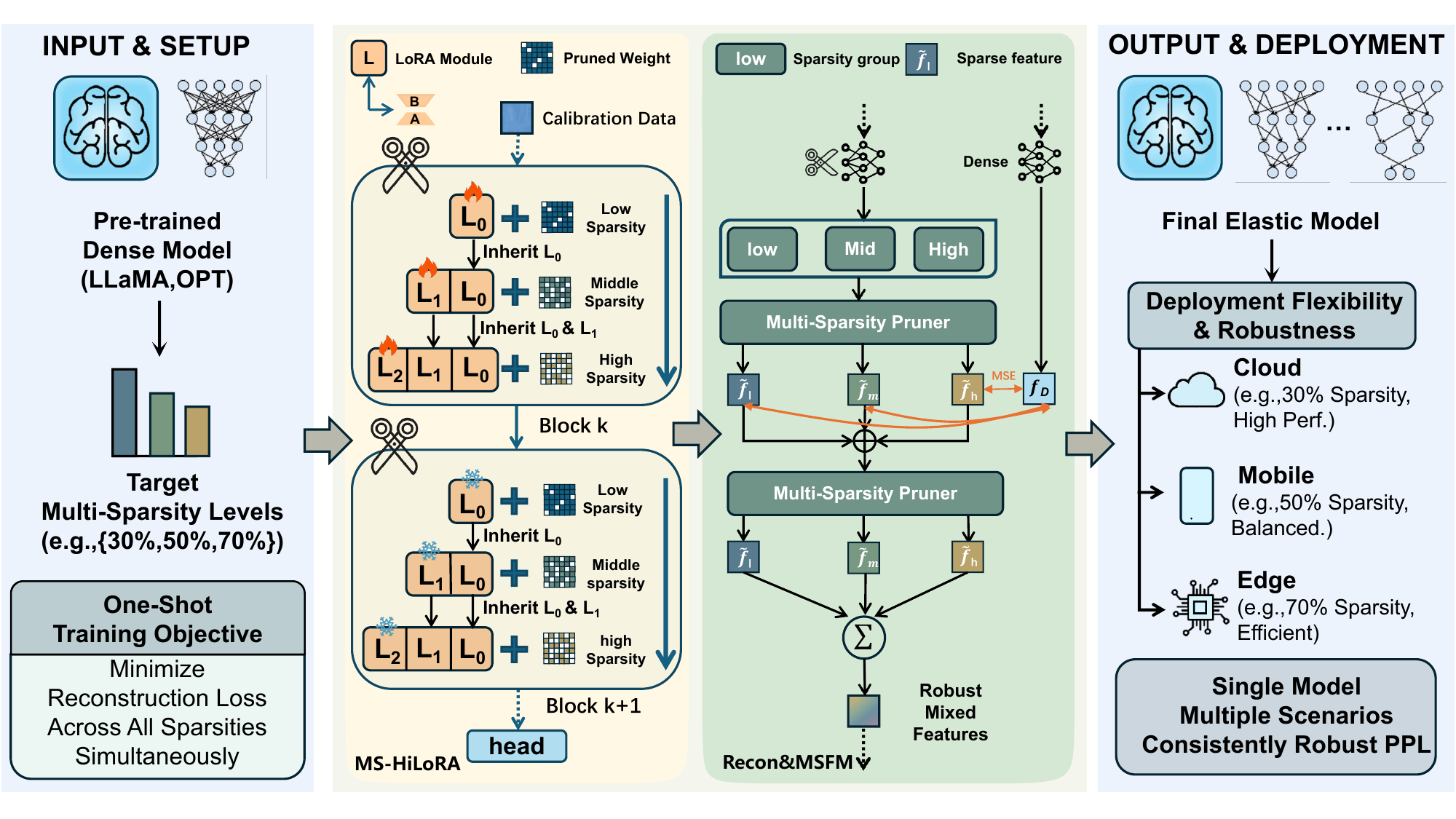}
    \caption{Overview of our proposed EPTS. Through block-wise reconstruction, EPTS compensates for performance degradation after pruning by fine-tuning LoRA modules L, while the original model weights remain frozen. The reconstruction process consists of two stages: (1) minimizing reconstruction loss across all sparsity groups simultaneously using Multi-Sparsity Hierarchy LoRA. (2) mixing multiple sparsity features across different sparsity groups by Multi-Sparsity Feature Mixer.}
    \label{general_figure}
\end{figure*}

To address the above limitation, we propose Elastic Post-Training Sparsity (EPTS), a unified framework that produces a versatile model capable of maintaining robust performance across diverse sparsity configurations through a single, one-shot reconstruction process. The overall pipeline is illustrated in the accompanying Figure~\ref{general_figure}. Specifically, we introduce a Multi-Sparsity Hierarchy LoRA (MS-HiLoRA) mechanism that establishes a parameter inheritance chain, enabling high-sparsity groups to reuse knowledge learned from lower-sparsity groups. Furthermore, we propose a Multi-Sparsity Feature Mixer (MSFM) between blocks to fuse features under different sparsity levels, thereby improving the robustness of the model to pruning rate perturbations during the block-wise reconstruction pass. Our contributions are summarized as follows:
\begin{itemize}
    \item We propose a novel EPTS framework built upon post-training sparsity, which enables obtaining a Multi-Sparsity elastic model that performs well across various sparsity levels with one-shot reconstruction.
    \item We devise the Multi-Sparsity Hierarchy LoRA (MS-HiLoRA) mechanism. By establishing a parameter inheritance chain, this mechanism facilitates effective knowledge transfer from low- to high-sparsity groups while mitigating the parameter competition inherent in multi-sparsity optimization.
    \item We introduce a Multi-Sparsity Feature Mixer, which effectively enhances the model’s adaptability to sparsity-induced perturbations by dynamically fusing feature representations under different sparsity granularities.
    \item Extensive experiments on the LLaMA and OPT families demonstrate that EPTS maintains competitive performance against state-of-the-art baselines. Crucially, it achieves superior deployment efficiency, allowing for instant sparsity switching to meet dynamic hardware constraints.
\end{itemize}
\section{Related Works}
\subsection{Post-Training Sparsity}
Post-Training Sparsity (PTS), drawing inspiration from Post
Training Quantization~\cite{DBLP:conf/eccv/FangSATGH20,DBLP:conf/icml/NagelABLB20}, has emerged as a resource-efficient paradigm for model pruning. Unlike conventional retraining-based approaches that require large-scale training with substantial computational overhead, PTS first zeroes out weights according to a predetermined criterion with layer-wise sparsity allocation, followed by lightweight compensation techniques to mitigate the resulting accuracy degradation. This paradigm requires only a small calibration dataset and significantly reduces the cost of obtaining sparse models. POT~\cite{POT} pioneers this paradigm and obtains sparse models through progressive layer-wise fine-tuning. Following this line, several subsequent works have further advanced the PTS paradigm. UniPTS~\cite{UniPTS} improves upon the conventional layer-wise Mean Squared Error (MSE) objective by designing a KL-divergence-based loss function with basis decay, and further integrates evolutionary search for layer-wise sparsity allocation with dynamic sparse training. FCPTS~\cite{FCPTS} associates pruning thresholds with sparsity ratios through a differentiable bridge function, enabling the learning of optimal per-layer sparsity allocation, while introducing a control loss to ensure that the global sparsity ratio precisely meets the target value. However, existing methodologies are typically confined to single-sparsity optimization, necessitating a separate optimization session for each target sparsity level. In contrast, our framework advances the PTS paradigm by addressing multi-sparsity optimization through a unified block-wise reconstruction approach.
\subsection{Low Rank Adaptation}
As one of the core techniques in parameter-efficient fine-tuning (PEFT), Low-Rank Adaptation (LoRA) has demonstrably reduced the cost of customizing large foundation models for downstream tasks by enabling lightweight yet high-performance adaptation across diverse domains. The key idea is to freeze the original pre-trained weights during fine-tuning and inject trainable low-rank decomposition matrices into the model to capture task-specific knowledge. This allows efficient adaptation with a minimal number of additional parameters while preserving the model's original capabilities. Following the LoRA framework, a series of LoRA variants~\cite{LoRA-FA,Delta-LoRA,Ada-LoRA,LoRA+} have been proposed. In the field of model quantization, QLoRA~\cite{DBLP:conf/nips/DettmersPHZ23} and QA-LoRA~\cite{DBLP:conf/iclr/XuXG0CZC0024} combine low-precision quantization with LoRA, further reducing storage and computational costs while maintaining competitive performance. However, in the context of model sparsity, existing works~\cite{LLM-Pruner} relegate LoRA to a post-hoc fine-tuning technique solely for performance recovery after pruning. Departing from this post-hoc paradigm, our work integrates hierarchical LoRA modules directly into the sparsification pipeline. This shift from passive recovery to active involvement enables the acquisition of sparse models at arbitrary sparsity levels within one-shot reconstruction, significantly bolstering efficiency.

\section{Methodology}

\subsection{Elastic Post-Training Sparsity Modeling}
In this section, we detail the modeling framework of Elastic Post-Training Sparsity (EPTS). The  objective of EPTS is to perform hierarchical reconstruction and compensation of model weights after initial pruning through an efficient optimization mechanism. This enables one-shot reconstruction to produce an elastic model that flexibly adapts to multiple sparsity configurations while preserving its original performance to the greatest extent possible.

We denote the pre-trained weight matrices of a large language model as {W}. Given a target sparsity set \(S\), where each \(s \in S\) represents a specific sparsity ratio, conventional post-training sparsity methods that adopt weight updating  require an independent, complete weight reconstruction process for each individual \(s\), which is highly inefficient. In contrast, EPTS aims to learn a set of universal, hierarchical LoRA modules through a single unified reconstruction process to compensate for the performance degradation caused by weight removal at different sparsity levels \(s\). Specifically, for a model composed of $L$ Transformer blocks, we perform reconstruction at the block-wise level. For the $l$-th block, let $W^l$ denote the original frozen weights of the internal linear layers. To obtain the sparse structure for a specific sparsity ratio $s$, we first generate a corresponding binary mask $M_s$. We adopt the activation-aware pruning metric proposed by Wanda \cite{Wanda} to evaluate weight importance:
\begin{equation}
     Score_{ij} = |W_{ij}| \cdot \|X_j\|_2,
\label{eq:wanda_score}
\end{equation}
where $\|X_j\|_2$ denotes the $L_2$ norm of the input feature $X_j$. Based on these scores, the binary mask $M_s$ is derived by determining a threshold to retain the most critical weights:
\begin{equation}
    M_{s} = \mathbbm{1}\left( \text{Score}_{ij} > \tau_s \right),
\label{mask}
\end{equation}
where $\mathbbm{1}(\cdot)$ is the indicator function and $\tau_s$ represents the score threshold satisfying the $s$-th percentile sparsity constraint.

The core idea of EPTS is to introduce a set of trainable hierarchical LoRA modules, bound to sparsity levels, to compensate for model degradation after pruning. We categorize the sparsity ratios into distinct groups, where each group is associated with a specific pair of low-rank matrices. In this framework, the reconstructed weight for a specific sparsity ratio $s$ is obtained by adding the corresponding hierarchically cumulative LoRA compensation to the original weight, followed by an element-wise multiplication with the mask $M_s$. Therefore, the optimization objective of EPTS can be formalized as minimizing the expected reconstruction error across all target sparsities:
\begin{equation}\min_{\{\Phi\}} \sum_{s \in \mathcal{S}} \mathbb{E}_{X \sim \mathcal{D}} \left[ \left( f^l(W; X) - f^l_s(\hat{W}_s | \Phi_s; X) \right)^2 \right],\label{eq:loss}
\end{equation}
where $\mathcal{D}$ denotes the calibration dataset, $f^l(\cdot)$ represents the dense output of the $l$-th block using the original pre-trained weights $W$. $f_s^l(\cdot)$ denotes the sparse output of the same block using the reconstructed weights $\hat{W}_s$ after pruning. Here, $\hat{W}_s = (W + \Phi_s) \odot M_s$ is the effective sparse weight matrix, where $\Phi_s$ represents the cumulative compensation parameters derived from the hierarchical modules for that specific sparsity. This mechanism ensures that higher sparsity levels can inherit the restoration capabilities learned by lower sparsity groups.
\subsection{The Pipeline of EPTS}
As shown in Figure \ref{general_figure}, the overall pipeline of EPTS operates through a unified block-wise reconstruction strategy designed to optimize for multiple sparsity levels simultaneously. Initially, pruning importance scores are computed for each layer within a block using an activation-aware metric prior to weight reconstruction, enabling the reuse of multi-sparsity output features in subsequent stages. To enable granular control without rigid numerical constraints, the global sparsity range is conceptually partitioned into distinct low, middle, and high intervals, each associated with specific sparsity-aware LoRA modules integrated into the linear layers. These modules are structured in a hierarchical inheritance mode, establishing cumulative parameter dependencies that facilitate knowledge transfer from lower to higher sparsity groups. During the elastic reconstruction process, sparsity ratios are uniformly sampled across these groups to generate block outputs under varying pruning conditions, after which the reconstruction losses between the sparse outputs and the original dense output are aggregated and backpropagated to jointly optimize the corresponding LoRA parameters. Following the optimization of the current block, a Multi-Sparsity Feature Mixer is employed to fuse the calibrated features from different sparsity levels, propagating a robust, mixed representation to the subsequent block to ensure performance stability across the entire target sparsity range. Upon completion, the model supports flexible deployment by simply applying a uniform sparsity across all layers and activating the corresponding LoRA modules to meet specific scenario requirements. The detailed reconstruction process is outlined in Algorithm 1 of Appendix A.

\subsection{Multi-Sparsity Hierarchy LoRA}
To address the inconsistent performance degradation observed across varying sparsity levels, we propose the \textit{Hierarchy LoRA mechanism}. Our approach is grounded in the \textit{Nested Information Loss Hypothesis}: we posit that under a consistent pruning metric, the information loss incurred at high sparsity levels inherently encompasses the loss observed at lower levels, compounded by additional structural degradation. Consequently, compensation parameters should theoretically exhibit a hierarchical dependency rather than independence.

Formally, we categorize the target sparsity ratios into $K$ distinct groups (e.g., $S_{low}, S_{mid}, S_{high}$ where $K=3$). We instantiate $K$ hierarchical LoRA modules, where the index $k \in \{0, 1, \dots, K-1\}$ represents the group index from low to high sparsity. Unlike standard LoRA which applies a single correction, we construct the compensation weights $\Phi_k$ as a cumulative summation of modules. The effective compensation increment for the $k$-th group is defined recursively:
\begin{equation}
\Phi_k = \underbrace{\Phi_{k-1}}_{\text {Inherited }}+\underbrace{B_k A_k}_{\text {Specific}}, \quad \text { with } \Phi_{0}=\mathbf{B_0 A_0},
\label{cumulative chain for BA}
\end{equation}
where $B_k$ and $A_k$ denote the low-rank matrices specifically assigned to the $k$-th sparsity group. Consequently, the sparse weights $\hat{W}_s$ for a specific sparsity ratio $s$ belonging to group $k$ are obtained by:
\begin{equation}
\hat{W}_s = (W + \Phi_k) \odot M_s = \left( W + \sum_{i=0}^{k} B_i A_i \right) \odot M_s.
\end{equation}
This formulation establishes a hierarchical inheritance mode: high-sparsity tasks (large $k$) implicitly inherit the feature restoration capabilities learned by lower sparsity levels (small $k$), while adding their own specific compensation ($B_k A_k$) to address severe pruning.

During reconstruction, we employ a progressive optimization strategy to minimize the reconstruction loss across all sparsity groups simultaneously. A uniform generator $P_k(s)$ samples sparsity rates from the defined groups, and the losses are aggregated. The joint optimization objective $\mathcal{L}_{total}$ is defined as:
\begin{equation}
\mathcal{L}_{total} = \sum_{k=0}^{K-1} \mathbb{E}_{s \sim P_k(s)} \left[ \mathcal{L}_{rec}(W, \hat{W}_s; X) \right],
\end{equation}
where $\mathcal{L}_{rec}$ represents the reconstruction loss between the dense output and the compensated sparse output.
\begin{equation}
\mathcal{L}_{rec}=\left\|W X-\hat{W}_sX\right\|_2^2.
\end{equation}

The core advantage of this hierarchical architecture lies in the asymmetric gradient flow it engenders. By analyzing the gradient flow with respect to the module parameters $\{B_k, A_k\}$, we derive the following relationship that governs the update mechanism:
\begin{equation}
    \frac{\partial \mathcal{L}_{\text{total}}}{\partial \{B_k, A_k\}} = \sum_{j=k}^{K-1} \frac{\partial \mathcal{L}_{rec}^{j}}{\partial (W+\sum_{i=0}^{j}B_iA_i)}.
\end{equation}
This formulation reveals a distinct asymmetric gradient distribution where the foundational parameters, denoted by index $k=0$, participate in the forward computation of every sparsity configuration $j \ge 0$, thereby accumulating aggregated gradients from all groups $j \in [0, K-1]$ during backpropagation . This comprehensive supervision signal compels the shared basis to capture the most robust and general feature representations, whereas the high-level parameters are activated exclusively for higher sparsity tasks, allowing them to focus specifically on refining information loss under aggressive pruning regimes . Ultimately, this hierarchical information learning mechanism secures a performance lower bound safeguarded by the base parameters while effectively mitigating parameter competition across disparate sparsity objectives.

\subsection{Multi-Sparsity Feature Mixer}
To establish an efficient and robust feature flow, we propose the Multi-Sparsity Feature Mixer (MSFM) module, which is designed to bolster model robustness by exploring feature representations at varying levels of sparsity granularity. We evaluated three fusion configurations:
\begin{itemize}
    \item Case 1: \textbf{Dense Passthrough}. In this baseline configuration, the output features from block $l$ bypass any sparsification process and are fed directly into block $l+1$ as dense outputs. This approach lacks the mechanism to propagate sparsity-aware features between blocks.
    \item Case 2: \textbf{Stochastic Substitution}. Features generated from different sparsity levels are randomly sampled to replace dense features and used as input for the subsequent reconstructed block $l+1$. While this introduces sparse information, its randomness leads to variations in the input. 
    \item Case 3: \textbf{Multi-Sparsity Feature Mixer}. This synthesizes a refined input for the subsequent reconstructed block $l+1$ by fusing outputs from multiple sparsity groups.
\end{itemize}
Experiments indicate that Case 3 achieves the best performance under high sparsity. Formally, for the $l$-th block, the input $X^{l+1}$ for the next block is constructed as a weighted linear aggregation of outputs, defined as:
\begin{equation}
    {X}^{l+1} = \sum_{k=0}^{K-1} \lambda_k \cdot {\left( \left( W^l + \Phi_k^l \right) \odot M_{s}^{k} \right) {X}^{l}}.
\end{equation}
where $\Phi_{k}^{l}$ denotes the cumulative LoRA compensation parameters for the $k$-th sparsity group, adhering to the hierarchical structure defined in Equation \ref{cumulative chain for BA}, while $\lambda_{k}$ acts as a balancing coefficient that modulates the contribution of each sparsity granularity to the final fused representation. This fused representation is subsequently propagated as the robust input for the following block. We prioritize this deterministic mixing strategy over stochastic substitution or simple dense pass-through methods because it explicitly integrates features across multiple sparsity levels, thereby extracting a stable consensus representation . By aggregating block output features across different sparsity levels, MSFM enhances the sensitivity of subsequent layers to varied sparsity patterns. Furthermore, this approach eliminates the randomness associated with stochastic methods, ensuring a smooth input representation that effectively mitigates the input distribution shifts typically observed in high-sparsity regimes.

\begin{table*}[!htb]
  \centering
  \scriptsize 
  \setlength{\tabcolsep}{3pt} 
  \resizebox{\textwidth}{!}{%
  \begin{tabular}{lccccccccccc}
    \toprule
    \multirow{2}{*}{\textbf{Method}} & \multirow{2}{*}{\textbf{Criterion}} & \multirow{2}{*}{\textbf{Sparsity}} & \multicolumn{2}{c}{\textbf{LLaMA-7B}} & \multicolumn{2}{c}{\textbf{LLaMA2-7B}} & \multicolumn{2}{c}{\textbf{LLaMA3-8B}} & \multicolumn{3}{c}{\textbf{OPT}(Wiki.($\downarrow$))}\\
    \cmidrule(lr){4-5} \cmidrule(lr){6-7} \cmidrule(lr){8-9} \cmidrule(lr){10-12}
    & & & Wiki.($\downarrow$) & 0-shot$^{7}$Avg.($\uparrow$) & Wiki.($\downarrow$) & 0-shot$^{7}$Avg.($\uparrow$) & Wiki.($\downarrow$) & 0-shot$^{7}$Avg.($\uparrow$) & 125M & 350M & 1.3B \\
    \midrule
    & & Dense & 5.68 & 64.24 & 5.47 & 64.21 & 6.13 & 68.23 & 27.65 & 22.00 & 14.62 \\
    \midrule

SparseGPT~\cite{SparseGPT} & \multirow{4}{*}{S-S} & \multirow{5}{*}{30\%} & 5.96 & \underline{63.55} & 5.77 & 63.12 & 6.69 & \textbf{67.78} & 28.83 & \textbf{23.19} & 15.16 \\
    Wanda~\cite{Wanda} & & & 5.99 & 63.14 & \underline{5.73} & \textbf{63.45} & \underline{6.67} & 67.10 & \textbf{28.10} & \underline{23.49} & \underline{15.00} \\
    RIA~\cite{RIA} & & & \textbf{5.86} & \textbf{63.60} & \textbf{5.64} & 63.42 & \textbf{6.60} & 67.06 & 28.63 & 23.54 & 15.16 \\
    ICP~\cite{ICP} & & & \textbf{--} & \textbf{--} & \textbf{--} & \textbf{--} & \textbf{--} & \textbf{--} & 28.66 & \textbf{--} & \textbf{14.67} \\
    \cmidrule{1-2} \cmidrule{4-12}
    \textbf{EPTS(Ours)} & M-S & & \underline{5.95} & {63.33} & 5.74 & \underline{63.37} & \underline{6.67} & \underline{67.11} & \underline{28.37} & 23.51 & 15.10 \\
    \midrule

    SparseGPT~\cite{SparseGPT} & \multirow{4}{*}{S-S} & \multirow{5}{*}{40\%} & \underline{6.32} & \underline{62.25} & 6.10 & 62.04 & 7.48 & \textbf{66.01} & \underline{30.48} & \textbf{25.80} & 16.50 \\
    Wanda~\cite{Wanda} & & & 6.38 & 61.90 & \underline{6.05} & \textbf{62.69} & 7.44 & 65.00 & 30.66 & \underline{26.54} & \underline{15.87} \\
    RIA~\cite{RIA} & & & \textbf{6.21} & 61.99 & \textbf{5.96} & \underline{62.62} & \textbf{7.36} & 64.87 & 30.50 & 26.55 & 15.91 \\
    ICP~\cite{ICP} & & & \textbf{--} & \textbf{--} & \textbf{--} & \textbf{--} & \textbf{--} & \textbf{--} & \textbf{29.66} & \textbf{--} & \textbf{15.04} \\
    \cmidrule{1-2} \cmidrule{4-12}
    \textbf{EPTS(Ours)} & M-S & & 6.33 & \textbf{62.65} & 6.07 & 62.37 & \underline{7.43} & \underline{65.38} & 30.74 & 26.65 & 16.03 \\
    \midrule

    SparseGPT~\cite{SparseGPT} & \multirow{4}{*}{S-S} & \multirow{5}{*}{50\%} & 7.19 & \underline{58.88} & 6.99 & \textbf{60.28} & \underline{9.42} & \textbf{63.02} & 36.90 & \underline{31.33} & \underline{17.45} \\
    Wanda~\cite{Wanda} & & & 7.26 & 58.27 & 6.92 & \underline{60.18} & 9.82 & 61.30 & 38.95 & 36.22 & 18.42 \\
    RIA~\cite{RIA} & & & \underline{7.13} & \textbf{59.93} & \underline{6.80} & 59.83 & \textbf{9.33} & \underline{62.18} & 37.61 & 36.64 & 18.05 \\
    ICP~\cite{ICP} & & & \textbf{--} & \textbf{--} & \textbf{--} & \textbf{--} & \textbf{--} & \textbf{--} & \textbf{33.90} & \textbf{--} & \textbf{16.49} \\
    \cmidrule{1-2} \cmidrule{4-12}
    \textbf{EPTS(Ours)} & M-S & & \textbf{6.99} & 57.90 & \textbf{6.86} & 58.62 & 9.48 & 60.26 & \underline{34.35} & \textbf{30.57} & 17.74 \\
    \midrule

    SparseGPT~\cite{SparseGPT} & \multirow{4}{*}{S-S} & \multirow{5}{*}{60\%} & \underline{10.18} & \underline{54.75} & \underline{10.13} & \textbf{55.21} & \underline{15.43} & \textbf{55.31} & 59.23 & \underline{50.84} & 22.17 \\
    Wanda~\cite{Wanda} & & & 10.73 & 53.97 & 10.77 & 53.95 & 23.52 & 49.58 & 75.15 & 90.55 & 26.49 \\
    RIA~\cite{RIA} & & & 10.63 & 52.28 & 10.40 & 54.14 & 19.64 & 52.40 & 70.26 & 98.53 & 26.28 \\
    ICP~\cite{ICP} & & & \textbf{--} & \textbf{--} & \textbf{--} & \textbf{--} & \textbf{--} & \textbf{--} & \underline{45.78} & \textbf{--} & \underline{21.17} \\
    \cmidrule{1-2} \cmidrule{4-12}
    \textbf{EPTS(Ours)} & M-S & & \textbf{8.64} & \textbf{55.35} & \textbf{8.52} & \underline{54.43} & \textbf{12.82} & \underline{54.22} & \textbf{42.70} & \textbf{39.65} & \textbf{20.94} \\
    \midrule

    SparseGPT~\cite{SparseGPT} & \multirow{4}{*}{S-S} & \multirow{5}{*}{70\%} & \underline{25.78} & \underline{45.05} & \underline{28.06} & \underline{44.40} & \underline{41.03} & \underline{44.90} & 223.53 & \underline{144.00} & 50.16 \\
    Wanda~\cite{Wanda} & & & 82.19 & 39.74 & 77.73 & 37.40 & 127.11 & 36.26 & 331.00 & 776.00 & 97.83 \\
    RIA~\cite{RIA} & & & 91.23 & 40.00 & 68.86 & 39.62 & 179.00 & 36.45 & 353.96 & 667.00 & 99.01 \\
    ICP~\cite{ICP} & & & \textbf{--} & \textbf{--} & \textbf{--} & \textbf{--} & \textbf{--} & \textbf{--} & \underline{65.20} & \textbf{--} & \underline{36.78} \\
    \cmidrule{1-2} \cmidrule{4-12}
    \textbf{EPTS(Ours)} & M-S & & \textbf{16.94} & \textbf{47.36} & \textbf{16.47} & \textbf{47.05} & \textbf{25.02} & \textbf{46.36} & \textbf{64.90} & \textbf{68.59} & \textbf{30.65} \\
    \bottomrule
  \end{tabular}%
  }
  \caption{Perplexity (PPL) on Wikitext2 and 0-shot Average Accuracy (0-shot$^{7}$Avg.) on seven tasks on LLaMA and OPT families. S-S means single-sparsity while M-S means multi-sparsity. The best results are highlighted in bold, while the second-best results are underlined.}
  \label{performance comparison}
\end{table*}

\section{Experimental Results}
\subsection{Experimental Setup}
While LLMs have proven highly effective across numerous domains, recent studies~\cite{jaiswal2024compressing} suggest that evaluating pruned models based solely on perplexity is inadequate. To address this, we implement a comprehensive evaluation framework in this section that integrates both standard perplexity metrics and a variety of sophisticated zero-shot NLP tasks. This multi-dimensional approach ensures a rigorous and holistic comparison of the models.

We conduct comprehensive evaluations with our method on the LLaMA\cite{DBLP:journals/corr/abs-2302-13971,Touvron2023Llama2O,DBLP:journals/corr/abs-2407-21783} and OPT families\cite{DBLP:journals/corr/abs-2205-01068}. We evaluate perplexity on the WikiText2 dataset and accuracy across seven zero-shot tasks: BoolQ\cite{DBLP:conf/naacl/ClarkLCK0T19}, RTE\cite{RTE}, HellaSwag\cite{DBLP:conf/acl/ZellersHBFC19}, WinoGrande\cite{DBLP:journals/cacm/SakaguchiBBC21}, ARC-Challenge\cite{DBLP:journals/corr/abs-1803-05457}, ARC-Easy\cite{DBLP:journals/corr/abs-1803-05457}, and OpenBookQA\cite{DBLP:conf/emnlp/MihaylovCKS18}. For LLM pruning baselines, we compare our method with SparseGPT\cite{SparseGPT}, Wanda \cite{Wanda}, and RIA\cite{RIA} and ICP\cite{ICP}. Consistent with standard setups, we employ 128 samples of 2048 tokens randomly selected from the C4 dataset for calibration and set the batchsize as 1 during the reconstruction process.

\subsection{Performance Comparison}
For LLMs, our framework demonstrates strong reconstruction capability. As shown in the Table \ref{performance comparison}, we benchmark EPTS against SparseGPT, a training-free method incorporating weight compensation, alongside Wanda and RIA, which are optimization-free approaches forgoing weight updates, and ICP, which employs a sliding-window-based block-level compensation strategy. In the low-sparsity regime (30\%–50\%), EPTS achieves performance comparable to highly efficient baselines. However, as the sparsity rate increases, the limitations of optimization-free methods become increasingly apparent. At high sparsity levels, methods lacking weight updates suffer from severe information loss. For instance, on LLaMA-7B, the perplexity of Wanda and RIA degrades significantly to 82.19 and 91.23 respectively under 70\% sparsity. In contrast, EPTS maintains a robust perplexity of 16.94, surpassing SparseGPT (25.78). This performance disparity widens further on OPT-125M, where the perplexity of SparseGPT, Wanda, and RIA collapses to 223.53, 331.00, and 353.96, respectively. Meanwhile, EPTS retains a much lower value of 64.90, demonstrating a clear advantage compared to ICP (65.20). Crucially, this trend is consistently observed across other models in the LLaMA and OPT families, where our method exhibits distinct superiority at 60\% and 70\% sparsity.

Furthermore, as presented in Table \ref{performance comparison}, EPTS exhibits appreciable advantages in zero-shot downstream task evaluations as the sparsity increases. For instance, at 70\% sparsity on the LLaMA-7B model, EPTS surpasses SparseGPT, Wanda, and RIA by average accuracy margins of 2.31\%, 7.62\%, and 7.36\% respectively across the seven evaluated tasks.

\subsection{Multi-Sparsity Hierarchy LoRA Ablation}
To investigate the efficacy of our Multi-Sparsity Hierarchy LoRA (MS-HiLoRA), we conducted an ablation study on LLaMA-7B and OPT-1.3B under three configurations: Case 1 (Independent Mode), where each sparsity group utilizes a separate LoRA module; Case 2 (Shared Mode), where a single LoRA module is shared across all sparsity groups; and Case 3 (MS-HiLoRA), where higher sparsity groups cumulatively activate and reuse LoRA modules from lower sparsity groups.

As shown in Table \ref{ablation for MS-HiLoRA}, the proposed Case 3 demonstrates consistent superiority across both model families, particularly in high-sparsity regimes. On LLaMA-7B: Case 3 consistently achieves the lowest perplexity. At the challenging 70\% sparsity level, Case 3 achieves a PPL of 16.94, outperforming Case 1 at 17.60 and significantly surpassing Case 2 at 22.01. This suggests that a simply shared LoRA (Case 2) struggles to resolve the parameter competition arising from diverse sparsity goals. On OPT-1.3B: The advantage of the hierarchical design is even more pronounced. While Case 1 and Case 2 suffer from catastrophic performance degradation at high sparsity levels with PPLs deteriorating to 82.29 and 91.22 at 70\% sparsity respectively, Case 3 maintains a robust PPL of 30.65. These results strongly validate the effectiveness of our hierarchical design. Unlike Case 1, which isolates learning and fails to transfer foundational knowledge, and Case 2, which faces optimization interference, MS-HiLoRA (Case 3) enables high-sparsity groups to inherit and refine foundational knowledge from lower levels.

To further analyze the impact of capacity allocation within the hierarchical structure, we investigated various LoRA rank distribution strategies across different sparsity groups based on OPT-1.3B. In our primary evaluations and the results reported for Case 3 in Table \ref{ablation for lora rank}, we used a default allocation strategy that assigns LoRA ranks of [24, 16, 16] to the low-, middle- and high-sparsity groups respectively. As illustrated in Table \ref{ablation for lora rank}, this configuration provides a robust performance baseline, notably achieving a PPL of 30.65 at 70\% sparsity. However, the results in Table \ref{ablation for lora rank} reveal that the performance of the EPTS framework can be further optimized by redistributing these trainable parameters: when the rank allocation is shifted to prioritize the high-sparsity group with the [16, 16, 24] configuration, the PPL at 70\% sparsity further improves to 30.27. This highlights that MS-HiLoRA is a highly flexible mechanism. By adjusting the rank allocation, we can "tilt" the model's accuracy curve to favor specific sparsity regimes, such as balanced cloud-based deployment or extreme edge-side compression without altering the overall unified optimization process.

\subsection{Multi-Sparsity Feature Mixer Ablation}
To validate the efficacy of the proposed Multi-Sparsity Feature Mixer (MSFM) module, we also conducted ablation studies on LLaMA-7B and OPT-1.3B architectures. We compared three different configurations illustrated in section 3.4. As shown in the Table \ref{ablation for MSFM}, experimental results demonstrate that Case 3 achieves superior performance compared to baseline schemes as the sparsity rate increases. From a mechanistic perspective: Unlike Case 1, where the lack of an MSFM module forces feature propagation between blocks to rely on dense output features, MSFM effectively aggregates block output features across different sparsity levels, enhancing the sensitivity of subsequent layers to different sparsity patterns. Compared to the stochastic strategy in Case 2, MSFM provides a deterministic and smooth latent representation for the subsequent reconstruction module, effectively mitigating input distribution shifts. For instance, in the low-sparsity level, Case 1 and Case 2 exhibit a marginal advantage over Case 3. However, as the sparsity increases, Case 3 (MSFM) demonstrates exceptional robustness. At 60\% sparsity, the advantage of MSFM begins to emerge. On LLaMA-7B, Case 3 achieves a PPL of 8.64, surpassing Case 1 (9.23) and Case 2 (8.87). Similarly, on OPT-1.3B, Case 3 records a PPL of 20.94, showing clear improvements over Case 1 (24.68) and Case 2 (22.21). The performance gap widens further at the extreme 70\% sparsity.  These results confirm that MSFM is critical to alleviate information loss induced by aggressive sparsification.
\begin{figure*}[!htb]  
    \centering
    \includegraphics[width=0.95\linewidth]{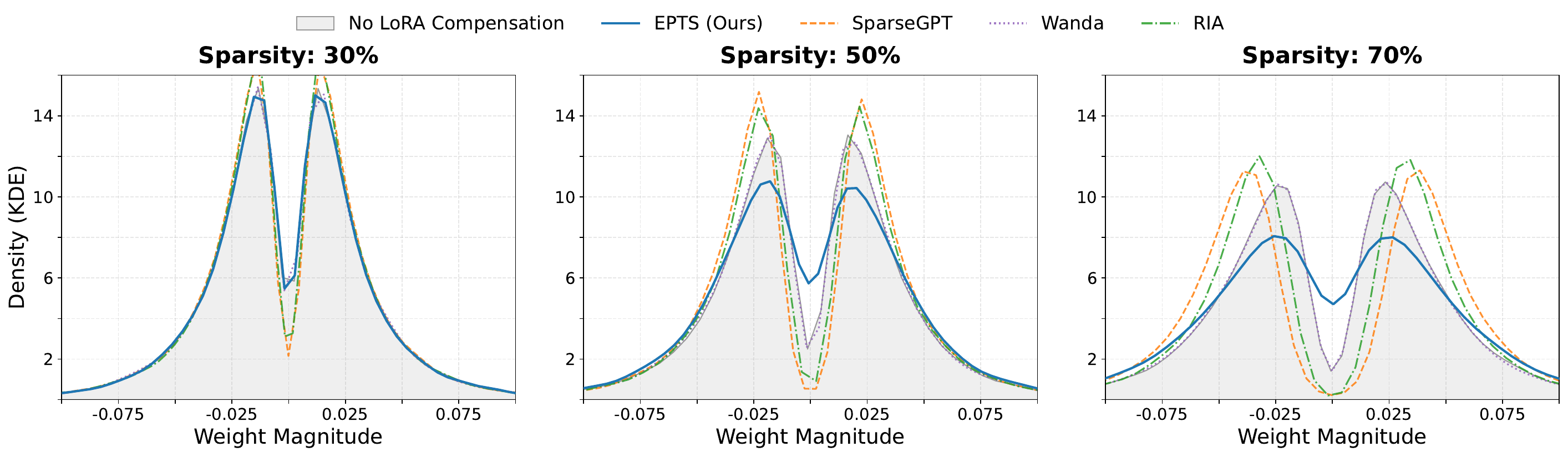}
    \caption{Kernel Density Estimation (KDE) visualization of weight distributions for the query projection layer (q\_proj) in the first Transformer block of LLaMA-7B. The plots compare the weight fidelity of EPTS against baselines (SparseGPT, Wanda, RIA) across 30\%, 50\%, and 70\% sparsity. }
    \label{weight_visual}
\end{figure*}

\begin{table}[!htb]
\centering
\resizebox{0.48\textwidth}{!}{%
\begin{tabular}{ccccc}
\toprule
\textbf{Model} & \textbf{Sparsity} & \textbf{Independent} & \textbf{Shared} & \textbf{MS-HiLoRA} \\
\midrule
\multirow{5}{*}{LLaMA-7B} & 30\% & 5.96 & 5.99 & \textbf{5.95} \\
 & 40\% & \textbf{6.33} & 6.37 & \textbf{6.33} \\
 & 50\% & 7.00 & 7.11 & \textbf{6.99} \\
 & 60\% & \textbf{8.64} & 8.80 & \textbf{8.64} \\
 & 70\% & 17.60 & 22.01 & \textbf{16.94} \\
\midrule
\multirow{5}{*}{OPT-1.3B} & 30\% & 15.10 & 15.10 & 15.10 \\
 & 40\% & 16.03 & 16.03 & 16.03 \\
 & 50\% & 22.26 & 18.56 & \textbf{17.74} \\
 & 60\% & 64.04 & 27.11 & \textbf{20.94} \\
 & 70\% & 82.29 & 91.22 & \textbf{30.65} \\
\bottomrule
\end{tabular}
 }
\caption{Ablation studies for Multi-Sparsity Hierarchy LoRA.}
\label{ablation for MS-HiLoRA}
\end{table}


\begin{table}[!htb]
  \centering
  \resizebox{0.48\textwidth}{!}{%
  \renewcommand{\arraystretch}{1.1}
  \setlength{\tabcolsep}{10pt}
  
  \begin{tabular}{cccc}
    \toprule
    \multirow{2}{*}{\textbf{Sparsity}} & \multicolumn{3}{c}{\textbf{Rank Configuration $\boldsymbol{[r_{low}, r_{mid}, r_{high}]}$}} \\
    \cmidrule(lr){2-4}
     & \textbf{[24, 16, 16]} & \textbf{[16, 24, 16]} & \textbf{[16, 16, 24]} \\
    \midrule
    50\% & 17.74 & \textbf{17.65} & 17.72 \\
    60\% & 20.94 & 21.04 & \textbf{20.88} \\
    70\% & 30.65 & 30.38 & \textbf{30.27} \\
    \bottomrule
  \end{tabular}
  }
  \caption{Ablation studies of LoRA rank allocation for Multi-Sparsity Hierarchy LoRA.}
  \label{ablation for lora rank}
\end{table}

\begin{table}[!htb]
\centering
\begin{tabular}{lcccc}
\toprule
\textbf{Model} & \textbf{Sparsity} & \textbf{Dense} & \textbf{Stochastic} & \textbf{MSFM} \\
\midrule
\multirow{5}{*}{LLaMA-7B} & 30\% & \textbf{5.89} & 5.94 & 5.95 \\
 & 40\% & \textbf{6.24} & 6.35 & 6.33 \\
 & 50\% & 6.99 & \textbf{6.95} & 6.99 \\
 & 60\% & 9.23 & 8.87 & \textbf{8.64} \\
 & 70\% & 25.48 & 19.21 & \textbf{16.94} \\
\midrule
\multirow{5}{*}{OPT-1.3B} & 30\% & 15.10 & 15.10 & 15.10 \\
 & 40\% & 16.03 & 16.03 & 16.03 \\
 & 50\% & \textbf{17.02} & 17.79 & 17.74 \\
 & 60\% & 24.68 & 22.21 & \textbf{20.94} \\
 & 70\% & 89.94 & 42.22 & \textbf{30.65} \\
\bottomrule
\end{tabular}
\caption{Ablation studies for Multi-Sparsity Feature Mixer.}
\label{ablation for MSFM}
\end{table}
\label{sec:msfm_sensitivity}


\begin{table}[!htb]
  \centering
  \resizebox{0.48\textwidth}{!}{%
    \renewcommand{\arraystretch}{1.1}
    \setlength{\tabcolsep}{10pt}
    
    \begin{tabular}{cccc}
      \toprule
      \multirow{2}{*}{\textbf{Sparsity}} & \multicolumn{3}{c}{\textbf{Fusion Ratio Configuration} $\boldsymbol{[\lambda_{low}, \lambda_{mid}, \lambda_{high}]}$} \\
      \cmidrule(lr){2-4}
       & \textbf{[0.4, 0.3, 0.3]} & \textbf{[0.3, 0.4, 0.3]} & \textbf{[0.3, 0.3, 0.4]} \\
      \midrule
      50\% & \textbf{17.37} & 17.41 & 17.74 \\
      60\% & 20.58 & \textbf{20.56} & 20.94 \\
      70\% & 31.58 & 31.27 & \textbf{30.65} \\
      \bottomrule
    \end{tabular}
  }
  \caption{Ablation studies of fusion ratio for Multi-Sparsity Feature Mixer.}
  \label{ablation for weight sum}
\end{table}

We further investigated the influence of the fusion weights $\lambda_{k}=[\lambda_{0},\lambda_{1},\lambda_{2}]$, which correspond to $[\lambda_{low}, \lambda_{mid}, \lambda_{high}]$, in the MSFM module on reconstruction quality based on OPT-1.3B. Consistent with the sparsity group defined in Section 3.3, we designed three different weight allocation strategies to isolate the impact of different feature granularities. As shown in Table \ref{ablation for weight sum}, each column represents a configuration where a dominant fusion weight ($\lambda=0.4$) is assigned to a specific target sparsity group ($S_{low}$, $S_{mid}$, or $S_{high}$), while the remaining groups are kept at a lower baseline of $\lambda=0.3$.

The experimental results demonstrate a precise correlation between the weight allocation and the performance of the corresponding sparsity group. Specifically, the optimal performance for each sparsity level is consistently achieved when its corresponding group is assigned the dominant weight: At 50\% sparsity within $S_{low}$, the strategy prioritizing $S_{low}$ achieves the lowest perplexity of 17.37. At 60\% sparsity within $S_{mid}$, the strategy prioritizing $S_{mid}$ yields the best result of 20.56. Crucially, at 70\% sparsity within $S_{high}$, the strategy prioritizing $S_{high}$ significantly outperforms the others, achieving a perplexity of 30.65 compared to 31.58 and 31.27 for the low- and mid-focused configurations respectively. This pattern confirms that increasing $\lambda_{k}$ explicitly guides the model to prioritize the optimization and feature utilization of the $k$-th sparsity group. Consequently, MSFM validates itself as a controllable mechanism, enabling the tailoring of model performance across different sparsity regimes simply by adjusting the fusion weights.

\subsection{Reconstruction Epochs Ablation}
\begin{figure}[!htb]
    \centering
    \includegraphics[width=1.0\linewidth]{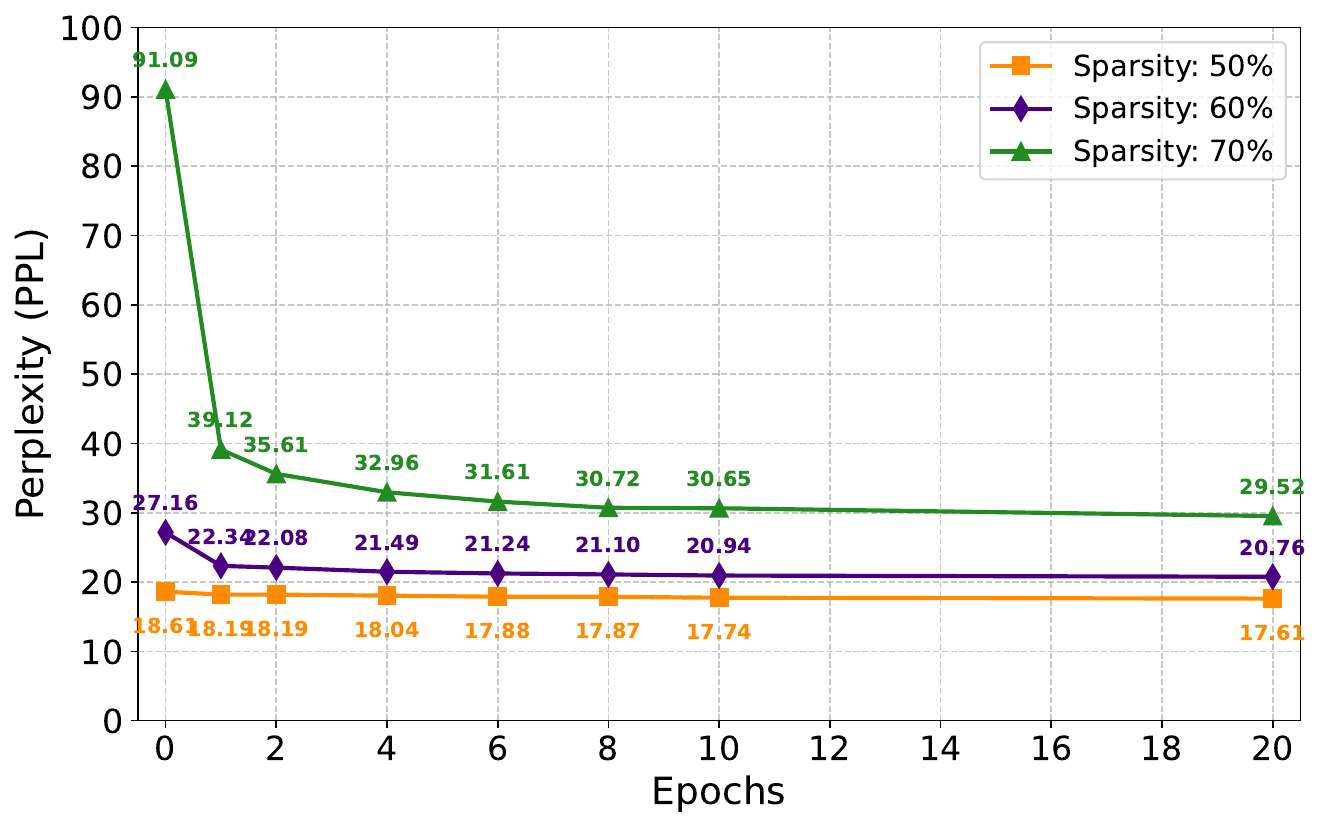}
    \caption{Ablation Study for Reconstruction Epochs.}
    \label{ablation for epoch}
\end{figure}

As shown in Figure~\ref{ablation for epoch}, under 70\% sparsity on OPT-1.3B, EPTS requires as few as 1 epoch to achieve substantial performance recovery, followed by rapid stabilization. Notably, for lower sparsity such as 50\% and 60\%, the model exhibits even faster convergence, attaining near-optimal performance with fewer optimization iterations compared to the 70\% setting. This characteristic allows us to flexibly choose the number of epochs to strike an optimal balance between performance and computational cost, thereby significantly reducing the actual time overhead while maintaining model performance. In our experiments, we set the number of epochs to 10.

\subsection{Inference Efficiency}
\begin{figure}[!htb]
    \centering
    \includegraphics[width=1\linewidth]{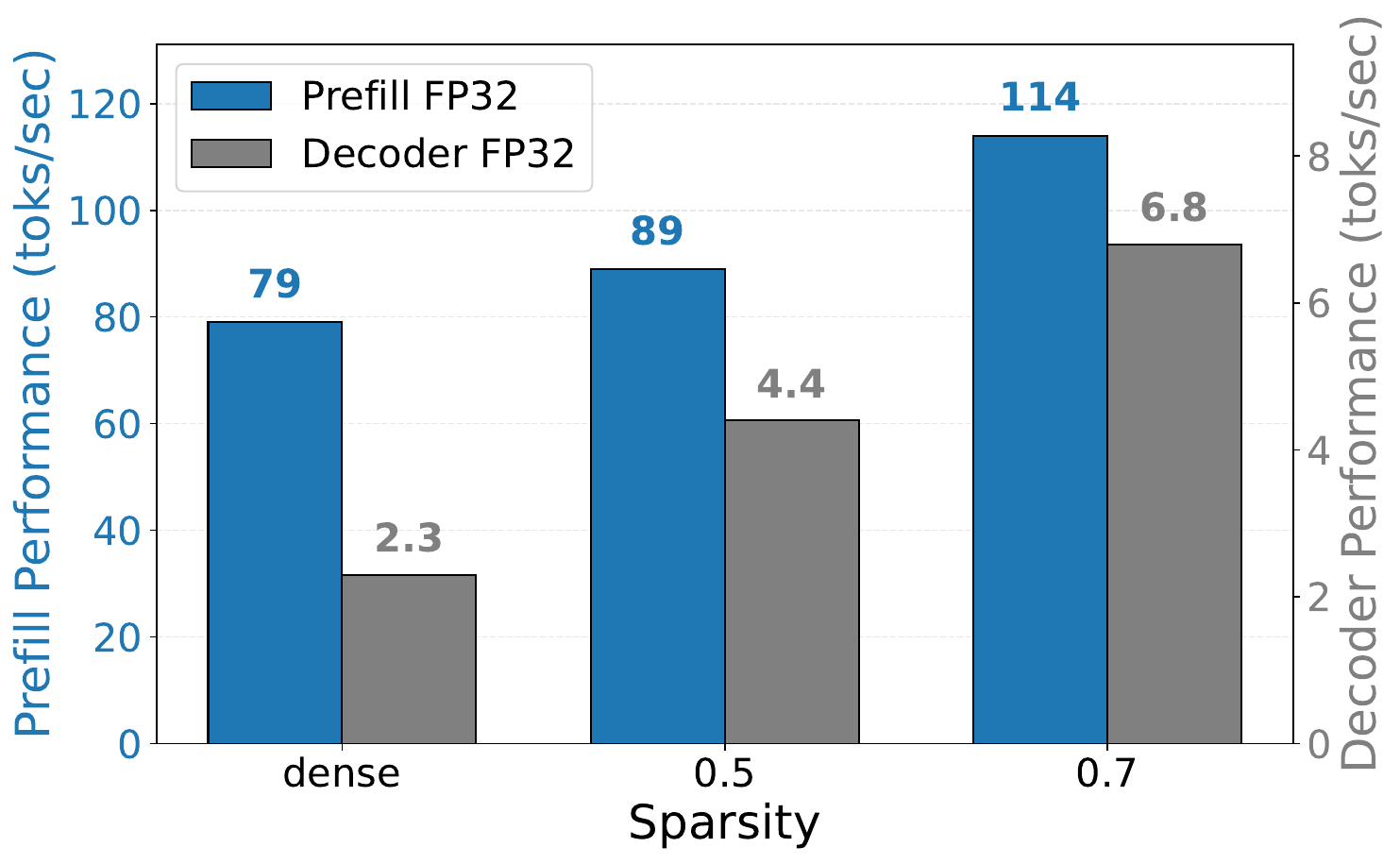}
    \caption{Inference Efficiency Comparison of LLaMA-7B on an 8-core AMD CPU (AWSc7a.4xlarge) using DeepSparse.}
    \label{speed}
\end{figure}

To present the efficiency of our method, we evaluated the LLama-7B model on an 8-core AMD CPU using DeepSparse. As shown in Figure~\ref{speed}, the prefill throughput remains approximately 89 tokens/s for the dense and 50\% sparsity models, increasing to 114 tokens/s at 70\% sparsity. For decode performance, the speed improves from 2.3 tokens/s for the dense model to 4.4 tokens/s at 50\% sparsity, reaching 6.8 tokens/s at 70\% sparsity. Notably, the reduced memory size through sparsity enabled an increase of approximately $2.96\times$ in decode tokens per second compared to the dense baseline.

\section{Analysis}
\subsection{Weight Distribution Fidelity}
To provide an intuitive explanation for the superior robustness of EPTS, particularly in high-sparsity regimes, we visualize the Kernel Density Estimation (KDE) of the weight distributions. We specifically focus on the query projection layer ({q\_proj}) within the first Transformer block (Layer 0) of the LLaMA-7B model. Figure \ref{weight_visual} presents a comparative analysis between No LoRA compensation after pruning, EPTS (Ours), and comparative methods (SparseGPT, Wanda and RIA) across sparsity levels of 30\%, 50\%, and 70\%. 

\noindent\textbf{Distribution Fidelity at High Sparsity.} The most significant divergence emerges at the 70\% sparsity level. As illustrated in the right panel of Figure \ref{weight_visual}, existing methods such as SparseGPT, Wanda, and RIA exhibit severe distribution distortion. Their weight densities are characterized by abnormally sharp peaks and narrowed variance. This indicates that, resulting from either reconstruction-based weight compensation as in SparseGPT or rigid metric-based selection as in Wanda and RIA, these methods confine the remaining parameters into unnatural numerical intervals, thereby compromising the natural statistical properties inherited from the pre-trained weights. In contrast, EPTS maintains a smooth and broad distribution curve, represented by the blue line. Leveraging the MS-HiLoRA mechanism, EPTS learns flexible low-rank residuals $\Phi_k$ rather than mechanically amplifying the remaining weights. This enables the model to compensate for information loss while preserving the intrinsic statistical distribution of the pre-trained weights.

\noindent\textbf{Consistency Across Sparsity Levels.} 
At low 30\% sparsity, the distributions across all methods exhibit considerable overlap, suggesting that standard pruning techniques are sufficient for mild compression. However, as sparsity increases to 50\% and 70\%, the "peaking" phenomenon in baseline methods intensifies, leading to the observed "weight distortion" in the probability density functions. EPTS exhibits consistent stability, effectively mitigating distribution shifts near zero-value boundaries. This visualization corroborates our \textit{Nested Information Loss Hypothesis} and confirms that EPTS successfully serves as a robust lower-bound protection mechanism against degradation induced by aggressive sparsification.

\section{Conclusion}
In this paper, we propose EPTS, a unified and efficient framework designed to overcome the limitations of existing post-training sparsity methods that are restricted to single-sparsity optimization. By introducing the MS-HiLoRA mechanism, EPTS cascades LoRA modules to alleviate parameter competition while enabling knowledge inheritance from low sparsity to higher sparsity groups. Furthermore, the proposed MSFM tangibly enhances the model's robustness against pruning perturbations by dynamically fusing feature representations of different sparsity granularities. Extensive experiments on mainstream LLMs demonstrate the superior performance of EPTS. However, our work still has limitations: the multi-sparsity collaborative optimization implemented by our method does not perform well at extremely high sparsity such as 80\% sparsity and above. In future work, we plan to integrate semi-structured and structured pruning to develop a comprehensive multi-granularity and multi-sparsity unified pruning framework.

\begin{acks}
This work was supported in part by the National Natural Science Foundation of China (No. 62576001, No. 62206003).
\end{acks}

\bibliographystyle{ACM-Reference-Format}
\bibliography{reference_simple}

\appendix

\section{Reconstruction Algorithm of EPTS}
To better present the pipeline of the EPTS reconstruction phase, we outline the detailed execution procedure in Algorithm \ref{algorithm}. This algorithm delineates the block-wise optimization strategy, systematically breaking down the process into two cooperative phases: the Multi-Sparsity Hierarchy LoRA (MS-HiLoRA) optimization for joint parameter learning, and the Multi-Sparsity Feature Mixer (MSFM) for robust feature fusion. By following these steps, EPTS effectively transforms a dense model into an elastic sparse model capable of adapting to diverse sparsity configurations within a single unified optimization session.
\begin{algorithm}[h]
\caption{EPTS: Block-wise Reconstruction with MS-HiLoRA and MSFM}
\label{algorithm}
\begin{algorithmic}[1]
\REQUIRE Pre-trained Weights $\{W^l\}_{l=0}^{L-1}$, Calibration Data $\mathcal{D}$, Sparsity Groups $\{\mathcal{S}_0, \mathcal{S}_1, \mathcal{S}_2\}$, Fusion Weights $\{\lambda_k\}_{k=0}^{K-1}$.
\ENSURE Compressed Model with Optimized Parameters.

\STATE \textbf{Initialization:}
\STATE $X_{in} \leftarrow \text{Emb}(\mathcal{D})$; \COMMENT{Get initial hidden states}

\FOR{block $l = 0$ to $L-1$}
    \STATE \textbf{Phase 1: MS-HiLoRA Optimization}
    \STATE Initialize learnable LoRA parameters $\{A_k, B_k\}_{k=0}^{K-1}$;
    \FOR{epoch $1 \dots E$}
        \FOR{each batch $x \in X_{in}$}
            \STATE $\mathcal{L}_{total} \leftarrow 0$; \textbf{optimizer.zero\_grad()}
            \FOR{group $k = 0$ to $K-1$}
                \STATE \textbf{1. Cumulative Compensation:}
                \STATE $\Phi_k = \sum_{j=0}^{k} B_j A_j$
                
                \STATE \textbf{2. Generate Sparsity Mask:}
                \STATE Sample sparsity $s \sim \mathcal{S}_k$ via uniform generator;
                \STATE Calculate importance scores $S = |W^l| \cdot \|x\|_2$;
                \STATE $M_s = \mathbbm{1}(S > \tau_s)$;
                
                \STATE \textbf{3. Forward Outputs:}
                \STATE $\hat{W}_s = (W^l + \Phi_k) \odot M_s$
                \STATE $\hat{Y}_s = \hat{W}_s x$; \quad $Y_{dense} = W^l x$
                
                \STATE \textbf{4. Reconstruction Loss ($\mathcal{L}_{rec}$):}
                \STATE $\mathcal{L}_{rec} = \| Y_{dense} - \hat{Y}_s \|_2^2$
                \STATE $\mathcal{L}_{total} \leftarrow \mathcal{L}_{total} + \mathcal{L}_{rec}$
            \ENDFOR
            \STATE \textbf{Backward:} $\nabla \leftarrow \partial \mathcal{L}_{total} / \partial \{A_k, B_k\}$; 
            \STATE \textbf{Update:} \textbf{optimizer.step()};
        \ENDFOR
    \ENDFOR

    \STATE \textbf{Phase 2: Multi-Sparsity Feature Mixer (MSFM)}
    \STATE Get optimized parameters $\{A_k^*, B_k^*\}$;
    \STATE Initialize output feature container $X^l \leftarrow \emptyset$;
    \FOR{each batch $x \in X^l$}
        \STATE \textbf{1. Multi-Granularity Forward:}
        \FOR{group $k = 0$ to $K-1$}
            \STATE $\Phi_k^* = \sum_{j=0}^{k} B_j^* A_j^*$;
            \STATE Sample $s \sim \mathcal{S}_k$ and get mask $M_s$;
            \STATE $y_k = ((W^l + \Phi_k^*) \odot M_s) x$;
        \ENDFOR
        
        \STATE \textbf{2. Weighted Fusion:}
        \STATE $y_{fused} = \sum_{k=0}^{K-1} \lambda_k \cdot y_k$
        \STATE Append $y_{fused}$ to $X^l$;
    \ENDFOR
    \STATE $X^{l+1} \leftarrow \text{Concatenate}(X^l)$; \COMMENT{Input for block $l+1$}
\ENDFOR
\end{algorithmic}
\end{algorithm}

\section{Complimentary Experiments}
\subsection{Ablation Studies}
To evaluate the data efficiency of EPTS, we conducted an ablation study on the LLaMA-7B model by varying the number of calibration samples($N$) from 1 to 128. The detailed results are presented in Table 6. For sparsity levels of 50\% and 60\%, the model's performance saturates rapidly with a minimal amount of data. Specifically, with only 32 samples, the model achieves perplexity scores of 17.86 at 50\% sparsity and 21.40 at 60\% sparsity, which are notably close to the results obtained with the full 128 samples (17.74 and 20.94, respectively). In the challenging 70\% sparsity scenario, while the model naturally requires more data to compensate for severe structural loss as indicated by the elevated PPL at $N=1$, it still demonstrates strong convergence properties. Performance improves significantly as $N$ increases, approaching an optimal plateau around 64 samples. At $N=64$, the PPL reaches 32.24, which is comparable to the result of 30.65 at $N=128$. This suggests that EPTS can effectively capture the necessary feature statistics for reconstruction with minimal data.
\begin{table}[htbp]
\centering
\begin{tabular}{cccccccc}
\toprule
\multirow{2}{*}{\textbf{Sparsity}} & \multicolumn{7}{c}{\textbf{Number of calibration samples } $\boldsymbol{N}$} \\
\cmidrule(lr){2-8}
 & \textbf{1} & \textbf{4} & \textbf{8} & \textbf{16} & \textbf{32} & \textbf{64} & \textbf{128} \\
\midrule
50\% & 18.87 & 18.00 & 18.03 & 18.07 & 17.86 & 17.81 & 17.74 \\
60\% & 26.43 & 23.24 & 22.42 & 22.02 & 21.40 & 21.06 & 20.94 \\
70\% & 70.40 & 50.30 & 42.09 & 38.21 & 34.87 & 32.24 & 30.65 \\
\bottomrule
\end{tabular}
\caption{Ablation study for the number of calibration samples}
\label{tab:nsamples_ablation}
\end{table}

\subsection{Time Efficiency}
We compare the computational time cost of different pruning methods on OPT-1.3B on NVIDIA RTX 4090 in Table \ref{runtime_comparison}.  Our proposed EPTS consumes relatively more time compared to these baselines. However, it is important to note that baselines like SparseGPT, Wanda, and ICP are target-specific (denoted by $\times N$), meaning they must be executed repeatedly ($N$ times) to obtain results for $N$ different sparsities. In contrast, EPTS employs a multi-sparsity optimization strategy, which allows it to optimize for a continuous range of sparsity levels simultaneously in a single run. Additionally, the runtime data for ICP is an approximate estimation derived from model scale scaling, as exact reproduction on the same hardware was not available.
\begin{table}[h]
\centering
\begin{tabular}{ccc}
\toprule
\textbf{Model} & \textbf{Method} & \textbf{Time (s)} \\
\midrule
\multirow{7}{*}{OPT-1.3B} & SparseGPT & $140 \times N$ \\
 & Wanda & $38 \times N$ \\
 & RIA & $46 \times N$ \\
 & ICP (1 epoch) & $176 \times N$ \\
 & ICP (12 epochs) & $510 \times N$ \\
 & EPTS (1 epoch) & 1201 \\
 & EPTS (10 epochs) & 8883 \\
\bottomrule
\end{tabular}
\caption{Time efficiency comparison on OPT-1.3B.}
\label{runtime_comparison}
\end{table}

\subsection{Block-wise Sequential Optimization}
While global optimization theoretically yields a global optimum, EPTS employs a sequential, block-wise reconstruction strategy to maximize practical efficiency. This sequential order naturally aligns with the network's forward-propagation dependency, allowing the Multi-Sparsity Feature Mixer (MSFM) to propagate stable consensus representations that mitigate input distribution shifts. Computationally, this approach restricts the peak memory footprint to a single block, preventing the Out-Of-Memory bottlenecks typical in global LLM optimization. Crucially, block-wise decoupling natively supports non-uniform sparsity deployment across layers without requiring any secondary training. To validate this, we performed a rapid optimal sparsity allocation search on OPT-1.3B combining KL-divergence sensitivity analysis (407s) and dynamic programming (0.9s). As demonstrated in Table \ref{non_uniform}, this non-uniform allocation consistently reduces perplexity compared to the strict uniform baseline across all sparsity thresholds.

\begin{table}[ht]
\centering
\begin{tabular}{ccc}
\hline
\textbf{Global Sparsity} & \textbf{Uniform} & \textbf{Non-Uniform} \\ \hline
30\% & 15.10 & \textbf{14.74} \\
40\% & 16.03 & \textbf{14.93} \\
50\% & 17.74 & \textbf{16.40} \\
60\% & 20.94 & \textbf{19.78} \\ \hline
\end{tabular}
\caption{Comparison of Uniform and Non-Uniform Sparsification on OPT-1.3B}
\label{non_uniform}
\end{table}

\begin{table}[ht]
\centering
\resizebox{\columnwidth}{!}{
\begin{tabular}{clccccc}
\hline
\textbf{Model} & \textbf{Method} & \textbf{30\%} & \textbf{40\%} & \textbf{50\%} & \textbf{60\%} & \textbf{70\%} \\ \hline
\multirow{2}{*}{{LLaMA-7B}} & Spec-LoRA & \textbf{5.87} & \textbf{6.21} & 7.02 & 9.35 & 28.14 \\
 & EPTS (Ours) & {5.95} & {6.33} & \textbf{6.99} & \textbf{8.64} & \textbf{16.94} \\ \hline
\multirow{2}{*}{{OPT-1.3B}} & Spec-LoRA & \textbf{14.68} & \textbf{15.18} & \textbf{16.77} & 21.51 & 37.31 \\
 & EPTS (Ours) & 15.10 & 16.03 & 17.74 & \textbf{20.94} & \textbf{30.65} \\ \hline
\end{tabular}
}
\caption{Effectiveness-Efficiency Trade-off Comparison}
\label{tradoff}
\end{table}

\subsection{Effectiveness-Efficiency Trade-off}
To evaluate whether unified optimization compromises recovery performance, we establish a Spec-LoRA baseline where an independent LoRA module is trained post-Wanda pruning for each target sparsity under an identical parameter budget. Table \ref{tradoff} details the results on LLaMA-7B and OPT-1.3B.

EPTS exhibits distinct superiority over Spec-LoRA in mid-to-high sparsity regimes. Under aggressive pruning at 70\% sparsity, independent training struggles severely to recover from massive structural weight loss. Conversely, the MS-HiLoRA mechanism forces a rigorous parameter inheritance chain that safeguards a robust performance lower bound. At 70\% sparsity on LLaMA-7B, the Spec-LoRA PPL degrades to 28.14, whereas EPTS maintains a superior PPL of 16.94.

At lower sparsity levels of 30\% and 40\%, the unified EPTS framework trails Spec-LoRA marginally under a strict uniform setting. Crucially, the decoupled EPTS architecture natively supports non-uniform layer-wise deployment without secondary optimization. Applying this non-uniform allocation on OPT-1.3B at 30\%, 40\%, and 50\% global sparsity thresholds reduces perplexity to 14.74, 14.93, and 16.40, respectively. This allows EPTS to directly outperform the sparsity-specific baselines at 40\% and 50\% sparsity, and achieve comparable performance at 30\%. Ultimately, EPTS eliminates hours of repetitive single-sparsity training without compromising competitive recovery.

\subsection{Scalability to Larger Models}
While our current empirical evaluations are constrained to models up to 8B parameters due to hardware limitations, the EPTS framework is inherently designed to scale effectively to massive architectures. Theoretically, our method benefits from the well-documented 'blessing of scale' in LLM compression. As observed in pioneering works such as SparseGPT~\cite{SparseGPT} and Wanda\cite{Wanda}, larger language models possess substantially higher parameter redundancy and intrinsic robustness to sparsification. Consequently, they suffer markedly less performance degradation than their smaller counterparts at identical sparsity levels.

Empirically, this trend is already evident within our current evaluations across the OPT and LLaMA families: the performance gap between the EPTS-pruned elastic models and their dense baselines steadily narrows as model capacity increases from 1.3B to 8B. Given that EPTS already achieves highly stable recovery on highly sensitive smaller architectures, the block-wise reconstruction mechanism is well-positioned to maintain similar effectiveness and elasticity when extended to larger-scale models.
\end{document}